\documentclass[10pt,twocolumn,letterpaper]{article}

\usepackage[pagenumbers]{cvpr} 

\usepackage[dvipsnames]{xcolor}

\definecolor{cvprblue}{rgb}{0.21,0.49,0.74}
\usepackage[pagebackref,breaklinks,colorlinks,citecolor=cvprblue]{hyperref}
\usepackage[accsupp]{axessibility} 

\usepackage{multirow}

\def\paperID{8836} 
\def\confName{CVPR}
\def\confYear{2024}

\begin{document}

\title{Anchor-based Robust Finetuning of Vision-Language Models}

\author{Jinwei Han$^{1}$, Zhiwen Lin$^2$, Zhongyisun Sun$^2$, Yingguo Gao$^2$, Ke Yan$^2$\\
Shouhong Ding$^2$, Yuan Gao$^{3\dag}$, Gui-Song Xia$^{1\dag}$\\
$^1$School of Computer Science, Wuhan University $^2$YouTu Lab, Tencent\\
$^3$Electronic Information School, Wuhan University\\
{\tt\small \{hanjinwei, guisong.xia\}@whu.edu.cn, \{sunzy12315, ethan.y.gao\}@gmail.com}\\
{\tt\small xavier.lin@foxmail.com, \{yingguogao, kerwinyan, ericshding\}@tencent.com}\\
}

\maketitle

\newcommand\blfootnote[1]{%
\begingroup
\renewcommand\thefootnote{}\footnote{#1}%
\addtocounter{footnote}{-1}%
\endgroup
}

\blfootnote{$^\dag$ Corresponding authors.}

\begin{abstract}

We aim at finetuning a vision-language model without hurting its out-of-distribution (OOD) generalization. We address two types of OOD generalization, i.e., i) domain shift such as natural to sketch images, and ii) zero-shot capability to recognize the category that was not contained in the finetune data. Arguably, the diminished OOD generalization after finetuning stems from the excessively simplified finetuning target, which only provides the class information, such as ``a photo of a \texttt{[CLASS]}''. This is distinct from the process in that CLIP was pretrained, where there is abundant text supervision with rich semantic information. Therefore, we propose to compensate for the finetune process using auxiliary supervision with rich semantic information, which acts as anchors to preserve the OOD generalization. Specifically, two types of anchors are elaborated in our method, including i) text-compensated anchor which uses the images from the finetune set but enriches the text supervision from a pretrained captioner, ii) image-text-pair anchor which is retrieved from the dataset similar to pretraining data of CLIP according to the downstream task, associating with the original CLIP text with rich semantics. Those anchors are utilized as auxiliary semantic information to maintain the original feature space of CLIP, thereby preserving the OOD generalization capabilities. Comprehensive experiments demonstrate that our method achieves in-distribution performance akin to conventional finetuning while attaining new state-of-the-art results on domain shift and zero-shot learning benchmarks.

\end{abstract}    

\vspace{-2.5mm}

\section{Introduction}
\label{sec:intro}

\begin{figure}[t]
\begin{center}
\resizebox{1\linewidth}{!}{
\includegraphics[trim={185pt 80pt 185pt 75pt},clip,width=1\linewidth]{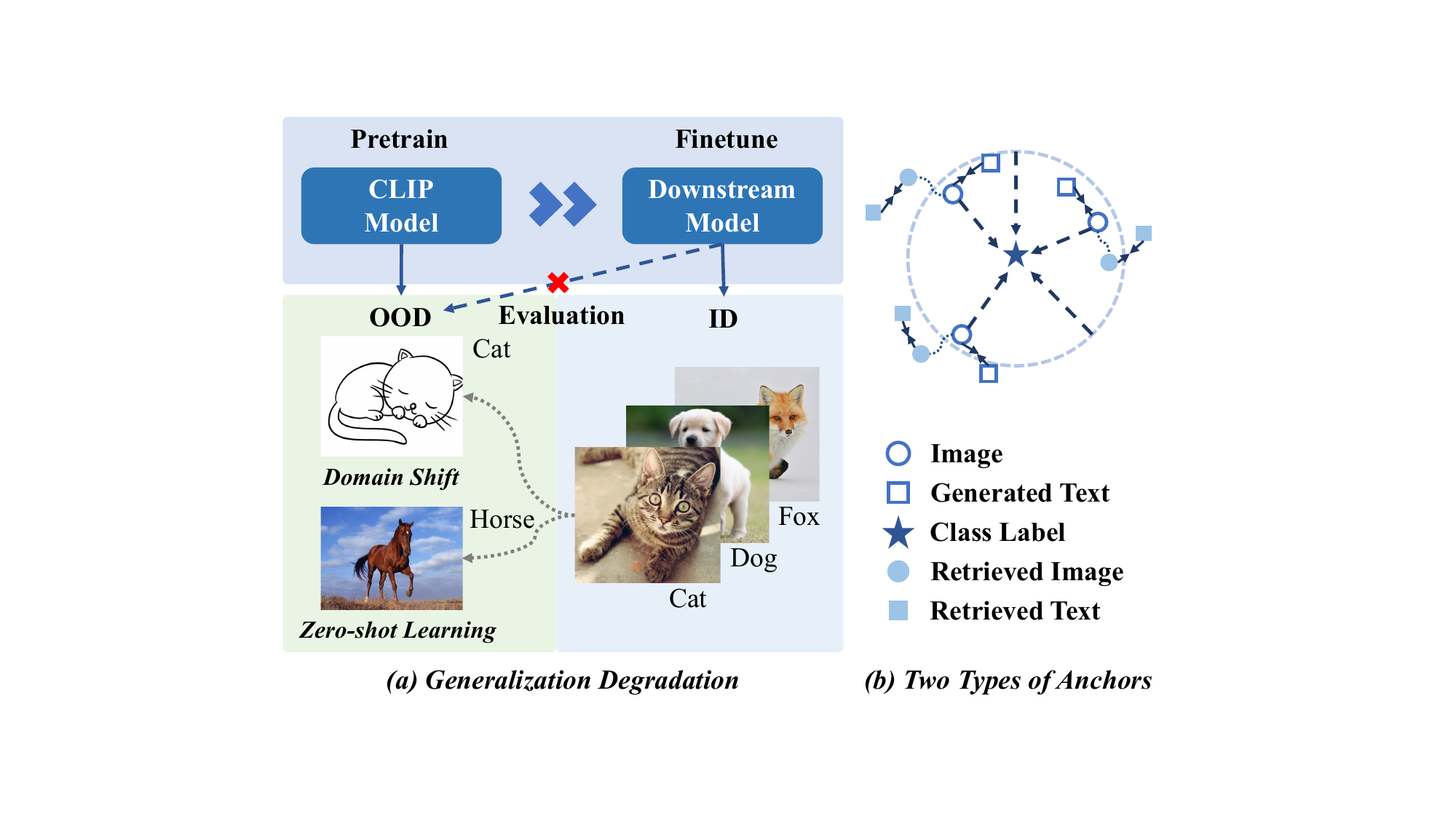}
}
\end{center}
\caption{Motivation illustration. (a) The out-of-distribution generalization (\textit{i.e.}, domain shift and zero-shot learning) capabilities of CLIP degrade significantly after finetuning on downstream tasks. (b) Images with generated texts and retrieved image-text pairs serve as two types of anchors to regularize the finetuning process of CLIP with auxiliary semantic information.}
\label{fig:intro}
\end{figure}

Maintaining out-of-distribution (OOD) generalization is essential for a pretrained model to ensure applicability across diverse circumstances (\textit{e.g.}, domain shift and zero-shot learning), even after adaptation to downstream tasks.
In recent years, contrastive vision-language pretrained models, such as CLIP \cite{CLIP} and ALIGN~\cite{ALIGN}, have exhibited exceptional OOD generalization capabilities in the aforementioned situations.
Although it is desirable to finetune these models using task-specific labeled data to improve performance on downstream tasks~\cite{CLIP-Finetuner, ViFi-CLIP}, they often experience a significant degradation of OOD generalization \cite{Wise-FT, LP-FT}. 

As illustrated in Fig. \ref{fig:intro}(a), OOD generalization encompasses domain shift, such as from natural to sketch images, and zero-shot capability to recognize categories not present in the finetuning data.
Numerous methods have been proposed to preserve the OOD generalization of CLIP during the finetuning process. 
Prompt learning \cite{CoOp, CoCoOp, remin} optimizes a set of learnable vectors using a limited number of labeled images while keeping the entire pretrained parameters of CLIP frozen.
Although these approaches demonstrate potential in addressing zero-shot prediction, their performance on downstream tasks is insufficient for practical needs.
Robust finetuning~\cite{Wise-FT, LP-FT, FLYP, TPGM} leverages all available data and employs a fully-tuned process to achieve high accuracy on in-distribution tasks without compromising performance under domain shift in OOD, while zero-shot capability remains neglected.
In this study, we extend robust finetuning to a more challenging scenario, aiming to preserve OOD generalization capabilities in both domain shift and zero-shot learning.
For instance, as shown in Fig. \ref{fig:intro}(a), our goal is to improve CLIP's capacity to recognize various types of animals while maintaining its original ability to classify animals from different domains (\textit{e.g.}, sketch) and other categories of animals (\textit{e.g.}, horse).

Our work is based on the observation that the feature space in CLIP, trained on large-scale image-text pairs, encompasses open-vocabulary semantic knowledge and thus demonstrates exceptional OOD generalization capabilities.
The conventional supervised finetuning paradigm primarily minimizes a cross-entropy loss on an image classifier with class labels.
FLYP \cite{FLYP} casts downstream class labels as text prompts (\textit{e.g.}, ``a photo of a \texttt{[CLASS]}'') and optimizes the contrastive loss to align the image embeddings with the prompt embeddings, using only the class information.

We argue that the decline in OOD generalization stems from the semantic-scarce supervision containing only class labels during the finetuning process.
Such an over-simplified finetuning target is distinct from the abundant text supervision employed in the pretraining of CLIP, leading to degraded OOD generalization.
Specifically, image features originally possessing rich semantics tend to collapse into a single class center after finetuning.
As a result, the feature space of the finetuned model shifts towards fitting the downstream dataset without the necessity of preserving rich semantics.
Finally, as a consequence, the original feature space of CLIP deteriorates significantly, impairing its OOD generalization capabilities.

Based on the above analysis, we propose an Anchor-based Robust Finetuning (ARF) approach that regularizes the finetuning process of CLIP with auxiliary contrastive supervision.
As illustrated in Fig. \ref{fig:intro}(b), our approach incorporates two types of anchors for maintaining the original feature space of CLIP.
Specifically, one of them is the text-compensated anchor, which is derived from using the image from the finetune set while enriching the semantic text from a pretrained captioner, such as BLIP2 \cite{BLIP2}. 
This is achieved by a carefully designed Text-Compensated Anchor Generation (TCAG) module. 
The other one is the image-text-pair anchor retrieved from a dataset similar to the pretraining data of CLIP according to the downstream task, whose text originally exhibits rich semantics. 
We implement this using the Image-Text Anchor Retrieval (ITAR) module.
These two types of image-text anchors are complementary to each other and are utilized as auxiliary supervision to regularize the finetuning process of CLIP, ensuring that the image features do not converge too close to the class prompt while retaining the original feature space of CLIP.
Extensive experiments demonstrate that our ARF achieves in-distribution performance comparable to finetuning while attaining new state-of-the-art OOD generalization results on domain shift and zero-shot learning benchmarks. 

Our main contributions are summarized as follows: 
\begin{itemize}
    \item We extend robust finetuning to a more challenging setting, aiming to preserve OOD generalization capabilities in both domain shift and zero-shot learning.
    \item We propose Anchor-based Robust Finetuning (ARF), using both text-compensated anchor and retrieved image-text-pair anchor to regularize the finetuning process.
    \item Extensive experiments reveal that our ARF attains new state-of-the-art performance on domain shift and zero-shot learning benchmarks while achieving in-distribution performance akin to finetuning.
\end{itemize}
\section{Related Work}
\label{sec:formatting}
\subsection{Vision-Language Contrastive Learning}
Vision-language models \cite{CLIP, ALIGN, FLAVA} primarily focus on establishing a joint embedding space for cross-modal learning by aligning web-scale images and texts. 
Recent advancements in the contrastive vision-language pretraining paradigm, particularly CLIP \cite{CLIP}, have demonstrated exceptional out-of-distribution (OOD) generalization capabilities across various downstream tasks (\textit{e.g.}, domain shift and zero-shot learning). 
Numerous subsequent studies have been proposed to further enhance contrastive image-text pretraining, such as masked language/image modeling \cite{MaskCLIP, KaimingCLIP}, hard sample mining \cite{Hard-Sample}, and retrieval augmentation~\cite{RA-CLIP, RECO}. 
BLIP \cite{BLIP} aims to improve image-text pretraining for unified vision-language understanding and generation, while BLIP2 \cite{BLIP2} incorporates Large Language Models (LLMs) to further improve generalization. 
Although these methods have indeed enhanced vision-language pretraining, finetuning remains necessary to improve performance on downstream tasks. 
A recent study~\cite{CLIP-Finetuner} demonstrates that CLIP attains superior or at least competitive performance on downstream tasks after finetuning, compared to conventional supervised pretraining models.
ViFi-CLIP \cite{ViFi-CLIP} implicitly models temporal cues and effectively finetunes image-level CLIP representations for videos.
However, the OOD generalization capabilities significantly degrade after finetuning, impairing the applicability of CLIP across diverse circumstances.

\subsection{Finetuning for Generalization}

Maintaining the OOD generalization capabilities of CLIP during the finetuning process has been extensively investigated through various approaches.
Prompt learning \cite{CoOp, CoCoOp, MaPLe} incorporates a small number of learnable prompt vectors to finetune CLIP, using a limited set of labeled images while keeping the pretrained model weights fixed.
Although these methods demonstrate potential in addressing zero-shot prediction, their performance on downstream tasks remains unsatisfactory.
Robust finetuning \cite{Wise-FT, LP-FT, FLYP, TPGM} leverages all available data and implements a fully-tuned process to achieve high accuracy on in-distribution tasks without sacrificing performance under domain shift in OOD.
Wise-FT \cite{Wise-FT} ensembles the weights of finetuned and original models, yielding remarkable gains in domain shift.
LP-FT \cite{LP-FT} initially trains the classification layer and then finetunes the entire network. 
This two-stage process substantially mitigates distortion of pretrained features and improves generalization. 
FLYP \cite{FLYP} demonstrates that a straightforward method of mimicking contrastive pretraining consistently outperforms finetuning approaches, while TPGM \cite{TPGM} automatically learns the constraint imposed on each layer for finetuning regularization. 
Nonetheless, the zero-shot learning capability is neglected by robust finetuning, leading to a substantial degradation of OOD generalization.
In light of these observations, we propose a more practical and challenging setting to finetune CLIP on downstream tasks without compromising both domain shift and zero-shot learning generalization capabilities.

\begin{figure*}
\begin{center}
\resizebox{0.95\linewidth}{!}{
\includegraphics[trim={30pt 90pt 30pt 90pt},clip,width=1\linewidth]{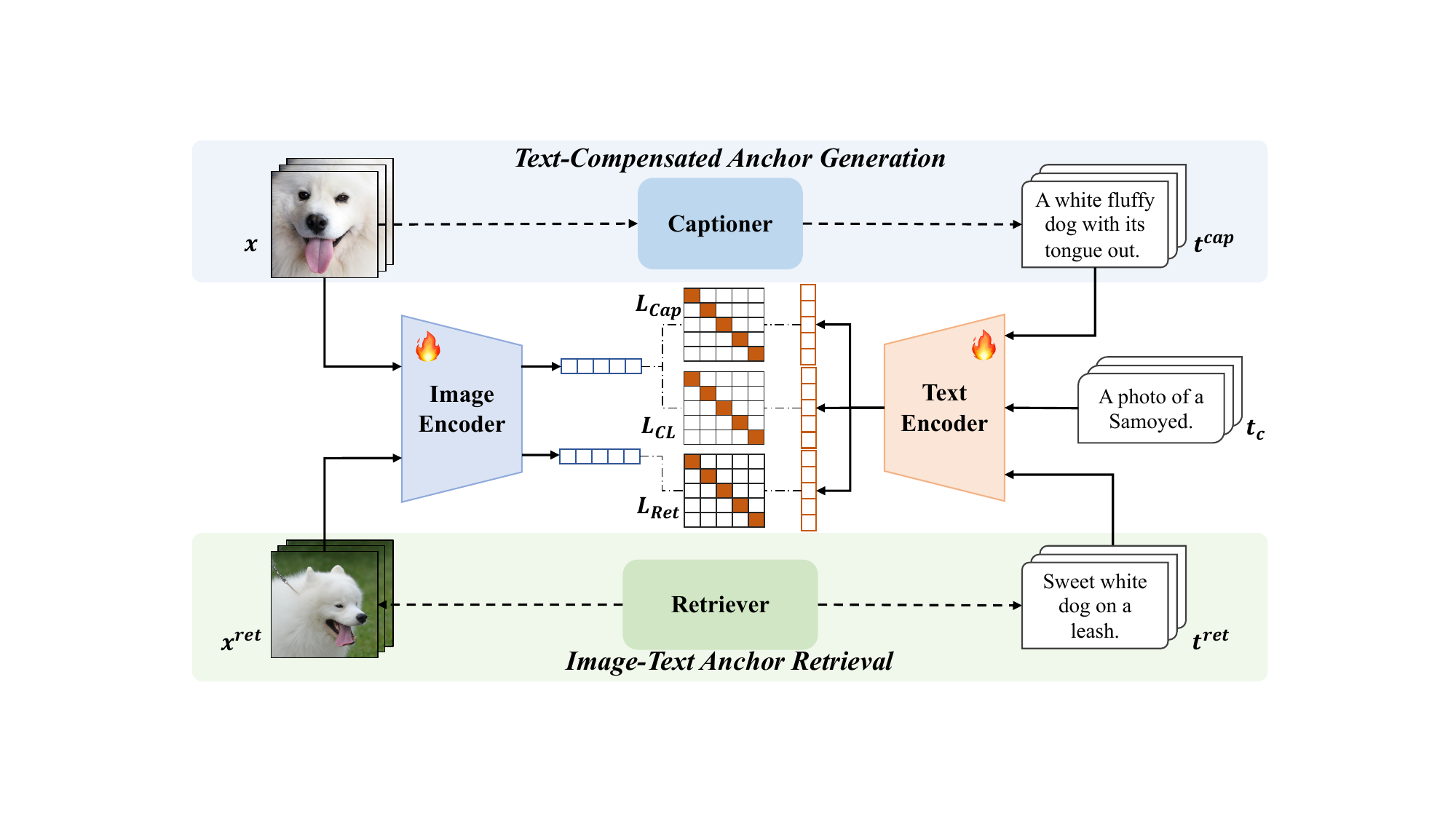}
}
\end{center}
   \caption{The pipeline of our proposed Anchor-based Robust Finetuning (ARF) comprises a Text-Compensated Anchor Generation (TCAG) module and an Image-Text Anchor Retrieval (ITAR) module. TCAG generates a caption for each image in the finetuning dataset utilizing a pretrained captioner as a text-compensated anchor with rich semantics. ITAR searches for image-text pairs from a candidate set similar to the data on which CLIP was pretrained, ensuring the presence of rich semantics in the image-text-pair anchor. We retrieve those samples related to our downstream tasks. A contrastive loss function, as used in CLIP, is employed for image-text alignment.}
\label{fig:framework}
\end{figure*}

\section{Preliminaries}

\subsection{Summary of Notations}
We summarize the notations used in the following sections. Specifically, we use the uppercase calligraphy font to denote a specific set, \eg, a dataset $\mathcal{S}$, which consists of input images and their corresponding class information. The input images are denoted by $\mathcal{X}$, while the class information is either represented by the class labels $\mathcal{Y}$, or the text descriptions $\mathcal{T}$ encompassing the class information, \eg, $\mathcal{S} = \{\mathcal{X}, \mathcal{Y}\}$ or $\mathcal{S} = \{\mathcal{X}, \mathcal{T}\}$. We denote the set of all the classes by $\mathcal{C}$. We use the lowercase font to denote a sample, \eg, $x$ is an image sample.

As for the scripts, we use superscript to indicate a specific data split, which can be a split of training (train), testing (test), in-distribution (id), domain shift (ds), zero-shot learning (zsl), etc. And the subscripts serve as indices. For example, $x_{i}^\text{train}$ is the $i$-th image sample from the training dataset where $x_{i}^\text{train} \in \mathcal{X}^\text{train}$.

\subsection{Contrastive Vision-Language Models}

In this paper, we adapt the contrastive vision-language pretraining model CLIP \cite{CLIP} to downstream tasks, which comprises an image encoder $f(\cdot)$ and a text encoder $g(\cdot)$. 
During the pretraining phase, image-text pairs are sampled from a web-scale training dataset $\mathcal{S}^{\text{train}}=\{(x_i^{\text{train}}, t_i^{\text{train}})\}$, where each text description $t_i^{\text{train}}$ encompasses rich semantics with abundant class information.
The categories $\mathcal{C}^{\text{train}}$ cover open-vocabulary semantic knowledge.
Each image $x_i^{\text{train}}$ and text $t_i^{\text{train}}$ is mapped to an image embedding $f(x_i^{\text{train}})$ and a text embedding $g(t_i^{\text{train}})$, respectively. 
Subsequently, a contrastive loss function $\mathcal{L}_{CL}$ is utilized to align the image embedding $f(x_i^{\text{train}})$ with the corresponding text embedding $g(t_i^{\text{train}})$, which is formulated as,
\begin{equation}
\begin{split}
\mathcal{L}_{CL} = -\frac{1}{B}\sum_{i=1}^B\log\frac{\exp(f(x_i^{\text{train}}) \cdot g(t_i^{\text{train}}) / \tau)}{\sum_{j=1}^B \exp(f(x_j^{\text{train}}) \cdot g(t_j^{\text{train}}) / \tau)} \\
-\frac{1}{B}\sum_{i=1}^B\log\frac{\exp(g(t_i^{\text{train}}) \cdot f(x_i^{\text{train}}) / \tau)}{\sum_{j=1}^B \exp(g(t_j^{\text{train}}) \cdot f(x_j^{\text{train}}) / \tau)}, 
\end{split}
\label{eq:1}
\end{equation}
where $B$ represents the number of image-text pairs in the minibatch, and $\tau$ denotes the temperature parameter used to scale the pairwise similarities in the loss function.

We consider image recognition as evaluation, that is, we evaluate the performance on a test set $\mathcal{S}^{\text{test}}=\{(x_i^{\text{test}}, y_i^{\text{test}})\}$ with categories $\mathcal{C}^{\text{test}}$.
CLIP utilizes predefined text prompts $t_{c}^{\text{test}}$ as inputs for the text encoder, which describe each class $c \in \mathcal{C}^{\text{test}}$, such as,
\begin{equation}
\text{text prompt:} \ a \ photo \ of \ a \ \texttt{[CLASS]}, 
\label{eq:2}
\end{equation}
where ``\texttt{[CLASS]}'' represents the class name. 
The output text embeddings $g(t^{\text{test}})$ are employed as classification weights during evaluation.
Given an image $x_{i}^{\text{test}}$ in $\mathcal{S}^{\text{test}}$, it is fed into the image encoder of CLIP to obtain the corresponding image embedding $f(x_{i}^{\text{test}})$. 
The predicted label $p_i^{\text{test}}$ is calculated as,
\begin{equation}
    p_i^{\text{test}}=\arg \max_{c \in {C^{\text{test}}}}(f(x_{i}^{\text{test}}) \cdot g(t_{c}^{\text{test}})),
\end{equation}
which implies that the class exhibiting the highest similarity between the image embedding $f(x_{i}^{\text{test}})$ and text embeddings $g(t_{c}^{\text{test}})$ of $\mathcal{C}^{\text{test}}$ classes is selected as the classification result.

We finetune the CLIP model by mimicking contrastive pretraining, as described in FLYP \cite{FLYP}.
Specifically, class labels are formulated as text prompts in Eq. \eqref{eq:2} and the contrastive loss function in Eq. \eqref{eq:1} is utilized to align image embeddings with text prompt embeddings.
We follow the same evaluation process as used in CLIP \cite{CLIP}.
\section{Method}
\subsection{Problem Setup}

Given a pretrained CLIP model, we aim to finetune it on the in-distribution dataset $\mathcal{S}^{\text{id}}=\{(x_i^{\text{id}}, y_i^{\text{id}})\}$ sampled from the distribution $P^{\text{id}}$ with classes $\mathcal{C}^{\text{id}}$, where each image $x_i^{\text{id}}$ has a label $y_i^{\text{id}} \in \mathcal{Y}^{\text{id}}$. 
The adapted model should perform at least as well as conventional finetuning methods on the test set from the same distribution $P^{\text{id}}$ and with the same categories $\mathcal{C}^{\text{id}}$ as the training data.
Simultaneously, we also work towards preserving the out-of-distribution (OOD) generalization capabilities in both domain shift and zero-shot learning scenarios.
In the domain shift situation, we evaluate the performance on the domain shift dataset $\mathcal{S}^{\text{ds}}=\{(x_i^{\text{ds}}, y_i^{\text{ds}})\}$.
The test data are sampled from a different domain $P^{\text{ds}}$ but share the same categories as the in-distribution data where we have $P(\mathcal{X}^{\text{id}}) \neq P(\mathcal{X}^{\text{ds}})$ while $P(\mathcal{Y}|\mathcal{X}^{\text{id}})=P(\mathcal{Y}|\mathcal{X}^{\text{ds}})$.
As for zero-shot learning, we have the downstream dataset as $\mathcal{S}^{\text{zsl}}=\{(x_i^{\text{zsl}}, y_i^{\text{zsl}})\}$, which composes of test image $x_i^{\text{zsl}}$ from a different category $y_i^{\text{zsl}} \in \mathcal{Y}^{\text{zsl}}$ where $\mathcal{C}^{\text{zsl}} \cap \mathcal{C}^{\text{id}} = \emptyset$.

In other words, as illustrated in Fig. \ref{fig:intro}(a), consider finetuning on real cat images, if we use sketch cat images for testing, then it is a domain shift problem; while if we test on horse images whose category is not included in the finetune set, then it is a zero-shot learning situation. We aim to achieve high performance on the in-distribution testing set, while also preserving the OOD generalization for both domain shift and zero-shot learning scenarios.

\subsection{Anchor-based Robust Finetuning}
\noindent\textbf{Overview.}
As shown in Fig. \ref{fig:framework}, our Anchor-based Robust Finetuning (ARF) approach finetunes both the image encoder and text encoder of CLIP using contrastive loss and incorporates two distinct modules to regularize the finetuning process. 
Specifically, {\em in the Text-Compensated Anchor Generation (TCAG) module}, we leverage a pretrained image captioner to generate a caption as text-compensated information for each image and align them using contrastive loss.
{\em In the Image-Text Anchor Retrieval (ITAR) module}, we search for image-text pairs from a dataset similar to CLIP's pretraining data, which are related to the downstream task.
These samples, originally containing rich semantic information, serve as additional anchors during finetuning.
These two types of anchors complement each other and preserve the original feature space of CLIP to ensure the OOD generalization after adaptation. 

\vspace{-2mm}
\subsubsection{Text-Compensated Anchor Generation}
\vspace{-1mm}
It is essential to note that merely employing the contrastive training loss between images $x$ and class prompts $t_c$ in Eq.~\eqref{eq:1} can lead to overfitting the excessively simplified finetuning target, as class prompts $t_c$ contain only class information shown in Eq. \eqref{eq:2}.
This semantic-scarce supervision is distinct from the abundant text supervision exploited in the pretraining of CLIP and can result in the degradation of the original feature space.

\noindent\textbf{Caption Generation.}
To mitigate the aforementioned issue, we propose to employ a pretrained image captioner, such as BLIP2~\cite{BLIP2}, to generate a text description $t_i^{\text{cap}}$ (\textit{i.e.}, caption) as compensated rich-semantic information for each image $x_i$, thereby preventing the overfitting on class prompts $t_c$.
The generated captions $t^{\text{cap}}$ encompass various descriptive words with more abundant semantics compared to class prompts $t_c$, similar to the texts utilized in CLIP's pretraining.
The images $x$ and corresponding captions $t^{\text{cap}}$ form the text-compensated anchors for regularization.

\noindent\textbf{Image-Caption Contrastive Learning.}
We cast $(x, t^{\text{cap}})$ as text-compensated anchors and utilize a contrastive loss function $\mathcal{L}_{\text{Cap}}$ similar to Eq. \eqref{eq:1} to align them within the feature space, thereby maintaining the semantic consistency between images and texts.
The text-compensated anchors prevent overfitting by ensuring that the conventional finetuning process in Eqs. \eqref{eq:1} and \eqref{eq:2} does not pull the embedding of the image $x_i$ too close to the text embedding of its corresponding class prompt $t_c$, regularizing the finetuning process of CLIP with auxiliary semantic supervision.


\begin{figure}
\begin{center}
\resizebox{1\linewidth}{!}{
\includegraphics[trim={180pt 120pt 170pt 110pt},clip,width=1\linewidth]{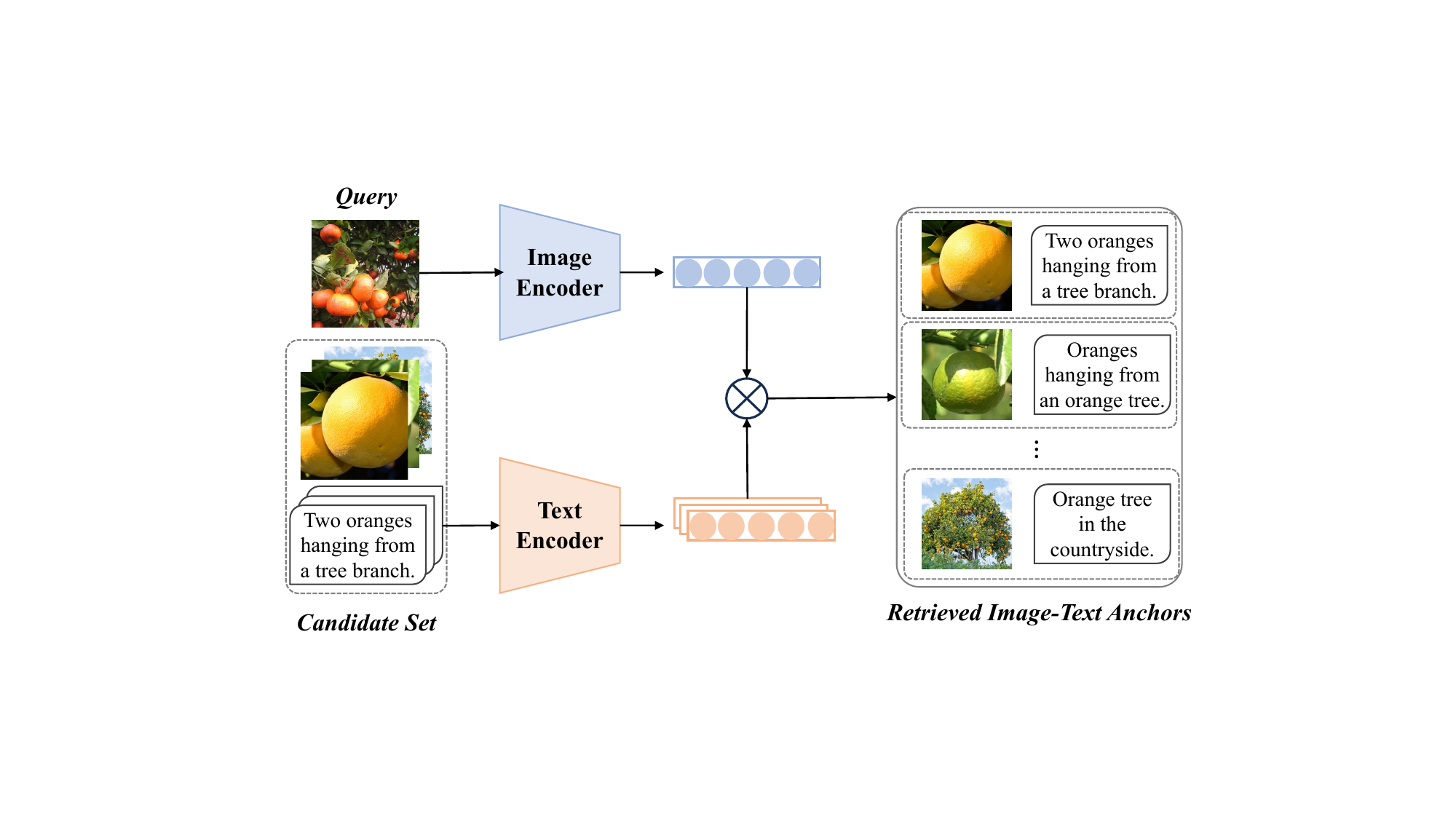}
}
\end{center}
   \caption{The pipeline of our Image-Text Anchor Retrieval (ITAR) module. We search for the most similar image-text pairs in the candidate set to obtain the rich semantic image-text anchors related to the downstream task for regularizing the finetuning process.}
\label{fig:retrieval}
\end{figure}

\subsubsection{Image-Text Anchor Retrieval}
Furthermore, we propose to search for image-text pairs according to the downstream task as auxiliary anchors.
We construct a candidate set that is similar to the pretraining data of CLIP, originally containing rich semantic information.
The pretrained CLIP model possesses an exceptional cross-modal retrieval capacity, which we employ to find image-text pairs with abundant semantics for regularization.
These retrieved image-text-pair anchors are aligned using a contrastive loss function to preserve the original feature space during the finetuning process. 

\noindent\textbf{Candidate Set Construction.}
We construct the web-scale image-text dataset CC3M \cite{CC3M} as a candidate set $\mathcal{S}^{\text{can}}=\{(x_i^{\text{can}}, t_i^{\text{can}})\}$, which is employed to search for rich-semantic image-text pairs.
The candidate set closely resembles the pretraining data of CLIP and encompasses abundant semantic information for maintaining the original feature space.
We leverage the cross-modal retrieval capacity of CLIP to obtain image-text-pair anchors relevant to the downstream task.
Specifically, we extract the embeddings of the images $x$ in the downstream dataset as $f(x)$, as well as the embeddings of the texts $t^{\text{can}}$ in the candidate set as $g(t^{\text{can}})$ for preparation.
These embeddings are utilized for retrieval and can be precomputed offline.

\noindent\textbf{Image-Text Pair Retrieval.}
In practice, only a small subset of the candidate set $\mathcal{S}^{\text{can}}$ is relevant to the downstream task and can be employed to preserve the original feature space of CLIP. We propose to search for image-text pairs from the candidate set $\mathcal{S}^{\text{can}}$ using KNN search.

Specifically, as illustrated in Fig. \ref{fig:retrieval}, we designate each image $x_i$ as the query and find the most similar image-text pairs from the candidate set $\mathcal{S}^{\text{can}}$ by calculating the similarity between the image embedding $f(x_i)$ and the text embeddings $g(t^{\text{can}})$, which can be formulated as follows,
\begin{equation}
    k = \arg \max(f(x_{i}) \cdot g(t^{\text{can}})),
\end{equation}
where $k$ represents the index of retrieved image-text pair in the candidate set $\mathcal{S}^{\text{can}}$.

The retrieval process can be efficiently executed using existing libraries, such as Faiss \cite{Faiss}.
Subsequently, we utilize the retrieved image-text pairs $\mathcal{S}^{ret}=\{(x_k^{ret}, t_k^{ret})\}$ with rich semantics as auxiliary anchors to regularize the finetuning process.

\noindent\textbf{Image-Text Contrastive Learning.}
We denote $(x^{ret}, t^{ret})$ as retrieved image-text-pair anchors and employ a contrastive loss function $\mathcal{L}_{Ret}$, similar to Eq. \eqref{eq:1}, to preserve the original feature space of CLIP.
The retrieved image-text pairs exhibit rich semantics related to the downstream task and serve as auxiliary supervision during finetuning.

These two types of image-text anchors with abundant semantic information are complementary to each other and are utilized as additional contrastive supervision to regularize the finetuning process of CLIP.
The image encoder and text encoder of CLIP are finetuned together with the following loss function,
\begin{equation}
\mathcal{L}=\mathcal{L}_{\textrm{\em CL}}+\mathcal{L}_{\textrm{\em Cap}}+\mathcal{L}_{\textrm{\em Ret}}.
\end{equation}

\begin{table*}[t]
\centering
\resizebox{0.99\textwidth}{!}{
\begin{tabular}{@{}cccccccccccccc@{}}
\toprule
& \multicolumn{7}{c|}{ImageNet} & \multicolumn{6}{c}{DomainNet}\\
\cmidrule{2-14}
Methods & \multicolumn{1}{c|}{ID} & Im-V2 & Im-R & Im-A & Im-Sketch & \multicolumn{1}{c|}{ObjectNet} & \multicolumn{1}{c|}{Avg. OOD} & \multicolumn{1}{c|}{ID} & Sketch & Painting & Infograph & \multicolumn{1}{c|}{Clipart} & Avg. OOD \\
\midrule
CLIP & \multicolumn{1}{c|}{68.3} & 61.9 & 77.7 & 50.0 & 48.3 & \multicolumn{1}{c|}{54.2} & \multicolumn{1}{c|}{58.4} & \multicolumn{1}{c|}{84.8} & 65.7 & 68.5 & 50.2 & \multicolumn{1}{c|}{72.1} & 64.1 \\
\midrule
LP & \multicolumn{1}{c|}{79.9} & 69.8 & 70.8 & 46.4 & 46.9 & \multicolumn{1}{c|}{50.4} & \multicolumn{1}{c|}{56.9} & \multicolumn{1}{c|}{86.3} & 57.4 & 61.5 & 45.6 & \multicolumn{1}{c|}{64.1} & 57.2 \\
FT & \multicolumn{1}{c|}{81.3} & 71.2 & 66.1 & 37.8 & 46.1 & \multicolumn{1}{c|}{51.6} & \multicolumn{1}{c|}{54.6} & \multicolumn{1}{c|}{89.5} & 61.8 & 65.6 & 49.0 & \multicolumn{1}{c|}{71.7} & 62.1 \\
LP-FT & \multicolumn{1}{c|}{81.7} & 72.1 & 73.5 & 47.6 & 50.3 & \multicolumn{1}{c|}{54.4} & \multicolumn{1}{c|}{59.6} & \multicolumn{1}{c|}{89.5} & 63.6 & 67.4 & 50.7 & \multicolumn{1}{c|}{73.4} & 63.8 \\
FLYP & \multicolumn{1}{c|}{82.6} & \textbf{73.0} & 71.4 & 48.1 & 49.6 & \multicolumn{1}{c|}{54.7} & \multicolumn{1}{c|}{59.4} & \multicolumn{1}{c|}{\textbf{89.8}} & 64.1 & 68.5 & 50.8 & \multicolumn{1}{c|}{74.0} & 64.3\\
\midrule
ARF & \multicolumn{1}{c|}{\textbf{82.7}} & 72.8 & \textbf{75.6} & \textbf{50.3} & \textbf{51.8} & \multicolumn{1}{c|}{\textbf{55.8}} & \multicolumn{1}{c|}{\textbf{61.3}} & \multicolumn{1}{c|}{\textbf{89.8}} & \textbf{65.3} & \textbf{69.5} & \textbf{51.1} & \multicolumn{1}{c|}{\textbf{74.9}} & \textbf{65.2} \\
\bottomrule
\end{tabular}
}
\caption{Domain shift results (\%) of state-of-the-art conventional finetuning and robust finetuning approaches on ImageNet and DomainNet benchmarks. The numbers represent the top-1 accuracy. We employ ImageNet and DomianNet-Real as the finetuning datasets, while the others serve as domain shift evaluation datasets. The best results are marked in \textbf{Black}.}
\label{tab:distribution_shift}
\end{table*}

\section{Experiments}
\noindent\textbf{Overview.}
We assess the effectiveness of our proposed Anchor-based Robust Finetuning (ARF) approach by comparing it with several baselines and providing implementation details for reproducibility. 
The evaluation of maintaining out-of-distribution (OOD) generalization capabilities is divided into two sections.
In Section \ref{sec:domain-shift}, we present results for domain shift, which was the focus of the original robust finetuning methods.
Subsequently, in Section \ref{sec:zero-shot}, we show results for our extended scenario (\textit{i.e.}, zero-shot learning).
Furthermore, we conduct an ablation study in Section \ref{sec:ablation} to evaluate the efficacy of our approach and showcase qualitative examples of two types of anchors in Section \ref{sec:qualitative examples}.

\noindent\textbf{Baselines.}
We compare our ARF with two conventional ways of finetuning pretrained models to downstream tasks using cross-entropy: linear probing (LP) and end-to-end fully finetuning (FT). Additionally, we investigate recent advancements of robust finetuning, such as LP-FT \cite{LP-FT}, which involves an initial linear probing followed by fully finetuning, and FLYP \cite{FLYP}, wherein finetuning is conducted in a pretraining-like manner.

\noindent\textbf{Implementation Details.}
We employ a batch size of 512 for finetuning on ImageNet \cite{ImageNet} and DomainNet \cite{DomainNet} with 10 epochs. 
We utilize a learning rate of $10^{-5}$ and a weight decay parameter of $0.1$.
ViT-B/16 \cite{ViT} is utilized as the image encoder of CLIP for finetuning.
Domain shift and zero-shot learning benchmarks are exclusively used for evaluation. 
We apply the same prompt templates as those employed in CLIP \cite{CLIP} and WiseFT \cite{Wise-FT} for training and inference.

\begin{figure}
\begin{center}
\resizebox{1\linewidth}{!}{
\includegraphics[trim={0pt 0pt 0pt 0pt},clip,width=1\linewidth]{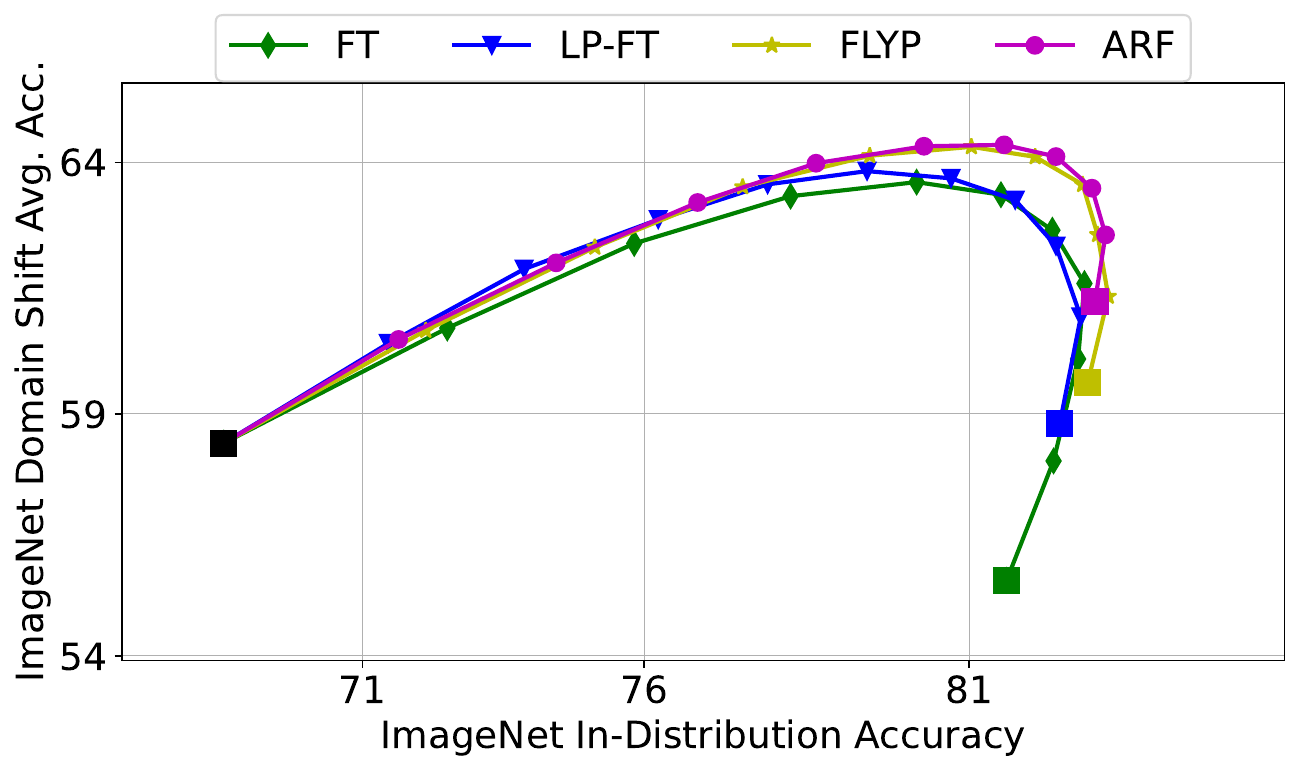}
}
\end{center}
   \caption{The ID and domain shift performance of our ARF compared with several baselines through linear interpolation of the finetuned model weights with the original model weights following Wise-FT \cite{Wise-FT}. The performance curves of ARF surpass (positioned in the upper right) those of the baselines on ImageNet, resulting in improved ID and domain shift accuracy.}
   \label{fig:ensemble}
\end{figure}

\subsection{Evaluation Under Domain Shift}
\label{sec:domain-shift}
\textbf{Benchmarks.}
We evaluate domain shift performance on two widely employed benchmarks, namely, ImageNet~\cite{ImageNet} and DomainNet \cite{DomainNet}. FMoW \cite{WILDS} and iWILDCam \cite{iwildcam} are excluded from our evaluation due to the poor domain shift performance of the original CLIP. 
In the first benchmark, we finetune CLIP using ImageNet \cite{ImageNet} for in-distribution evaluation and assess its performance on five distinct variants of ImageNet with domain shift: ImageNet-V2 \cite{ImageNetV2}, ImageNet-Sketch~\cite{ImageNet-Sketch}, ImageNet-A~\cite{ImageNet-A}, ImageNet-R \cite{ImageNet-R} and ObjectNet~\cite{ObjectNet}. We follow the training protocol outlined in FLYP \cite{FLYP} and employ the same ImageNet prompt templates used by CLIP during inference. 
For the second benchmark, we utilize the standard domain shift dataset DomainNet \cite{DomainNet} for evaluation. We finetune CLIP using DomainNet-Real for in-distribution performance and assess its generalization ability to four domain shift splits: Clipart, Infograph, Painting, and Sketch.

\noindent\textbf{Quantitative Results.}
We compare the performance of our ARF with several baselines on domain shift benchmarks, as detailed in Table \ref{tab:distribution_shift}.
On the ImageNet dataset, our ARF exhibits a slight performance advantage over conventional finetuning methods and other robust finetuning methods in the in-distribution (ID) test dataset.
The true strength of our ARF lies in its capacity to generalize to domain shift scenarios, achieving state-of-the-art performance with an average accuracy of 61.3\% across five domain shift test datasets.
Notably, our ARF performs significantly better than other finetuning methods, approaching or even surpassing the performance of CLIP on ImageNet-R, ImageNet-A, ImageNet-Sketch, and ObjectNet, which have large domain differences.
In the case of DomainNet, we also observe that our ARF demonstrates state-of-the-art performance of 65.2\% on the domain shift test dataset without sacrificing accuracy on in-distribution (ID) data. 
Our ARF outperforms CLIP on domain shift scenarios by 1.1\%, whereas other finetuning methods either fail to surpass it or achieve only marginal improvements.
These results indicate the effectiveness of the two types of anchors in our ARF. They effectively prevent the overfitting of the original feature space of CLIP to the downstream class prompts after finetuning, preserving the OOD generalization capability of CLIP for handling domain shift scenarios. 


\begin{table*}[t]
\centering
\resizebox{0.99\textwidth}{!}{
\begin{tabular}{@{}ccccccccccccc@{}}
\toprule
& \multicolumn{12}{c}{Zero-Shot Learning}\\
\cmidrule{2-13}
Methods & \multicolumn{1}{c|}{ImageNet} & Caltech & Flowers & Food & SUN & DTD & Aircraft & Cars & Pets & EuroSAT & \multicolumn{1}{c|}{UCF} & Avg. OOD\\
\midrule
CLIP & \multicolumn{1}{c|}{68.3} & 89.3 & 70.4 & 89.2 & 65.2 & 46.0 & 27.1 & 65.6 & 88.9 & 54.1 & \multicolumn{1}{c|}{69.8} & 66.6\\
\midrule
FT & \multicolumn{1}{c|}{81.3} & 78.8 & 16.0 & 37.3 & 39.3 & 29.7 & 4.7 & 10.8 & 80.2 & 15.4 & \multicolumn{1}{c|}{44.3} & 35.7\\
LP-FT & \multicolumn{1}{c|}{81.7} & 84.0 & 44.3 & 68.8 & 49.9 & 37.9 & \textbf{15.8} & 37.7 & 81.9 & 30.4 & \multicolumn{1}{c|}{59.5} & 51.0\\
FLYP & \multicolumn{1}{c|}{82.6} & 87.6 & 36.8 & 62.8 & 52.0 & 36.9 & 8.7 & 31.1 & 77.6 & 34.3 & \multicolumn{1}{c|}{58.6} & 48.6\\
\midrule
ARF & \multicolumn{1}{c|}{\textbf{82.7}} & \textbf{88.6} & \textbf{46.4} & \textbf{74.5} & \textbf{63.8} & \textbf{40.4} & 13.9 & \textbf{44.7} & \textbf{83.1} & \textbf{35.8} & \multicolumn{1}{c|}{\textbf{64.6}} & \textbf{55.6}\\
\bottomrule
\end{tabular}
}
\caption{Zero-shot learning results (\%) of state-of-the-art conventional finetuning and robust finetuning methods on numerous recognition tasks. The numbers represent the top-1 accuracy. We employ ImageNet as the finetuning dataset while the others serve as zero-shot learning evaluation datasets. The best results are marked in \textbf{Black}.}
\label{tab:zero-shot}
\end{table*}

\noindent\textbf{Weight Ensembling Curves.}
Wise-FT \cite{Wise-FT} demonstrates that a simple linear interpolation between the weights of the pretrained and finetuned models yields the optimal performance for both ID and domain shift. 
Therefore, we compare the performance of our ARF with the baselines by interpolating their model weights using 10 mixing coefficients ranging from 0 to 1.
As depicted in Fig. \ref{fig:ensemble}, we can see that our ARF outperforms the baselines after finetuning on ImageNet, leading to enhanced ID and domain shift accuracy.
Concretely, when comparing the coefficient that achieves the highest ID performance, our ARF with weight ensembling improves domain shift accuracy by 1.2\% over the state-of-the-art method (\textit{i.e.}, FLYP \cite{FLYP}). Comprehensive results are provided in the supplementary material.

\begin{table}[t]
\begin{center}
\resizebox{0.86\linewidth}{!}{
\begin{tabular}{cc|cc|c}
\toprule
\multicolumn{2}{c|}{Method}&\multicolumn{2}{c|}{ImageNet}&\multicolumn{1}{c}{Zero-Shot}\\
TCAG&ITAR&ID&Domain Shift&Avg. Acc\\
\midrule
\multicolumn{2}{c|}{baseline}&82.6 &59.4 & 48.6 \\
\midrule
 \checkmark & &82.6 &60.7(+1.3) & 54.3(+5.7) \\
 &\checkmark &82.6 &60.2(+0.8) & 53.6(+5.0) \\
 \checkmark &\checkmark &82.7 &61.3(+1.9) & 55.6(+7.0) \\
\bottomrule
\end{tabular}
}
\end{center}
\caption{Ablation study for Text-Compensated Anchor Generation (TCAG) module and Image-Text Anchor Retrieval (ITAR) module of our ARF. The baseline only conducts visual-language contrastive learning for finetuning like FLYP \cite{FLYP}.}
\label{tab:ablation}
\end{table}

\subsection{Evaluation Under Zero-shot Learning}
\label{sec:zero-shot}
\textbf{Benchmarks.}
We evaluate zero-shot learning performance using a diverse benchmark that encompasses a range of recognition tasks. 
To ensure a fair comparison, we employ the standard test split for inference, and the CLIP model is finetuned on ImageNet \cite{ImageNet}.
As for the evaluation of zero-shot learning, we utilize fine-grained object classification tasks such as OxfordPets \cite{OxfordPets}, StanfordCars \cite{StanfordCars}, Flowers102 \cite{Flowers102} and Food101 \cite{Food101}; as well as specific recognition tasks such as UCF101 \cite{UCF101} for action recognition, FGVCAircraft \cite{FGVCAircraft} for aircraft classification, DTD \cite{DTD} for texture classification, SUN397 \cite{SUN397} for scene recognition and EuroSAT \cite{EuroSAT} for satellite image classification. Additionally, we also evaluate our ARF on the general object classification dataset Caltech101 \cite{Caltech101}. 

\noindent\textbf{Quantitative Results.}
We present the results of our ARF and several baseline approaches on various zero-shot learning recognition tasks, as displayed in Table \ref{tab:zero-shot}. It can be observed that previous finetuning methods exhibit substantial improvements in accuracy on the ImageNet test data after finetuning with ImageNet training data, compared to the original CLIP. However, their performance significantly deteriorates in zero-shot learning recognition on the categories that were not contained in the finetuning data. In contrast, our ARF optimally maintains the zero-shot recognition capability without compromising the performance on ImageNet. The experimental results demonstrate that our ARF effectively regularizes the finetuning process with auxiliary semantic supervision, preserving the OOD generalization capability of CLIP for handling zero-shot learning scenarios. 

\begin{figure*}
\begin{center}
\resizebox{0.85\linewidth}{!}{
\includegraphics[trim={80pt 130pt 80pt 120pt},clip,width=1\linewidth]{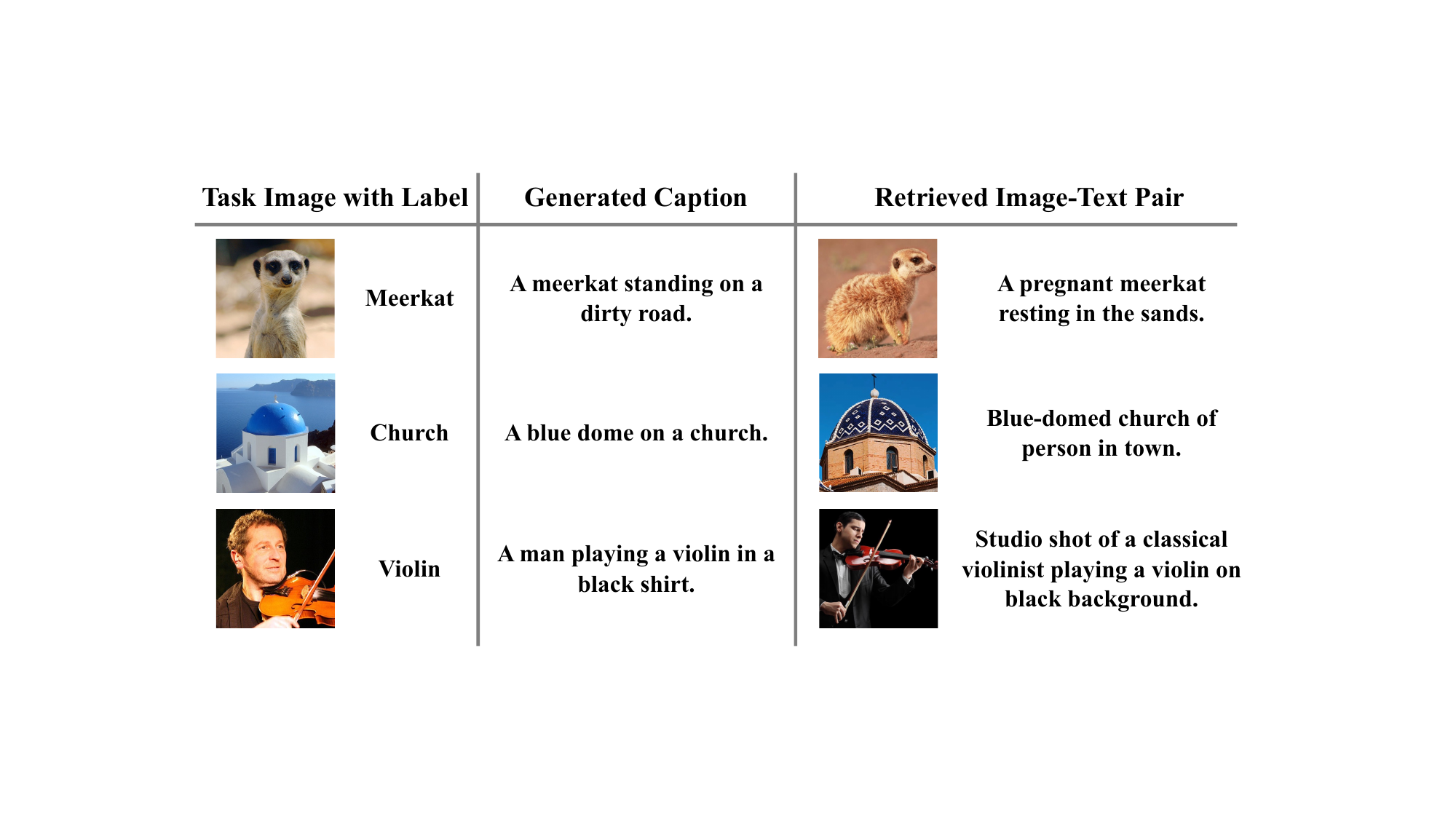}
}
\end{center}
\caption{Visualization examples of captions generated by a pretrained captioner (\textit{e.g.}, BLIP2 \cite{BLIP2}) and retrieved image-text pairs relative to the downstream task from the CC3M dataset \cite{CC3M}. The downstream task images and generated captions serve as text-compensated anchors for regularization. The retrieved image-text pairs function as auxiliary anchors for maintaining the feature space.}
\label{fig:example}
\end{figure*}

\subsection{Ablation Study}
\label{sec:ablation}
\noindent\textbf{Two Types of Anchors.}
To evaluate the effectiveness of our ARF, we conduct an ablation study to analyze the influence of the two types of anchors, as shown in Table~\ref{tab:ablation}. It can be observed that the Text-Compensated Anchor Generation (TCAG) module significantly improves domain shift and zero-shot learning accuracy by $1.3\%$ and $5.7\%$ over the baseline, respectively. 
These gains demonstrate that the rich semantic text descriptions generated by the pretrained captioner provide effective auxiliary supervision for the images to alleviate overfitting on class prompts. 
For retrieved image-text-pair anchors, we evaluate the influence of the Image-Text Anchor Retrieval (ITAR) module, which can improve domain shift and zero-shot learning accuracy over the baseline by $0.8\%$ and $5.0\%$ after finetuning on ImageNet, respectively. 
These results reveal that the rich semantic image-text pairs, retrieved from the candidate set similar to the pretraining data of CLIP according to the downstream task, are beneficial for regularizing the finetuning process. 
The two modules, working together, boost performance over the baseline by $1.9\%$ on domain shift and $7.0\%$ on zero-shot learning. 
The experimental results suggest that these two types of anchors complement each other and are beneficial to preserving the OOD generalization capabilities of CLIP.

\begin{table}[t]
\begin{center}
\resizebox{0.98\linewidth}{!}{
\begin{tabular}{c|cc|c}
\toprule
\multirow{2}{*}{Captioner}&\multicolumn{2}{c|}{ImageNet}&\multicolumn{1}{c}{Zero-Shot}\\
&ID&Domain Shift&Avg. Acc\\
\midrule
 BLIP \cite{BLIP} &82.1 &60.1 &54.0 \\
 BLIP2 \cite{BLIP2} &82.7 &61.3 & 55.6 \\
 BLIP2 + Vicuna \cite{vicuna} &82.5 &61.2 &56.5 \\
\bottomrule
\end{tabular}
}
\end{center}
\caption{Ablation study for the quality of generated captions. We evaluate the effectiveness of pretrained image captioners (\textit{i.e.}, BLIP \cite{BLIP} and BLIP2 \cite{BLIP2}) and further rewrite the text descriptions using Large Language Models (\textit{e.g.}, Vicuna \cite{vicuna}) in our ARF.}
\label{tab:captioner}
\end{table}

\noindent\textbf{The Quality of Captions.}
To assess the impact of caption quality in our ARF, we examine two pretrained image captioners (\textit{i.e.}, BLIP \cite{BLIP} and BLIP2 \cite{BLIP2}) and further rewrite the text descriptions using Large Language Models (\textit{e.g.}, Vicuna \cite{vicuna}).
As illustrated in Table \ref{tab:captioner}, employing captions generated by BLIP2 results in a 1.2\% improvement in domain shift performance and a 1.6\% enhancement in zero-shot learning performance compared to using captions generated by BLIP.
These gains demonstrate that more accurate text descriptions with rich semantics are beneficial for regularizing the finetuning process.
Additionally, we utilize Vicuna \cite{vicuna} to rewrite the captions for increased diversity and richer semantic information.
Since the text descriptions generated by BLIP2 are already sufficiently accurate, there is no improvement in ID and domain shift scenarios.
However, the rich semantic knowledge accessed from Vicuna boosts zero-shot learning by 0.9\%.
These results indicate the effectiveness of auxiliary information from LLMs, which warrants further exploration.

\subsection{Qualitative Examples of Anchors}
\label{sec:qualitative examples}
In Fig. \ref{fig:example}, we provide visualization examples to facilitate an understanding of how our Anchor-based Robust Finetuning (ARF) works.
The text descriptions (\textit{i.e.}, captions) generated by a pretrained captioner (\textit{e.g.}, BLIP2 \cite{BLIP2}) accurately describe the images, thus serving as effective anchors that supply rich information for maintaining the semantic consistency between images and texts.
The image-text pairs, retrieved from the candidate set, bear a close resemblance to the pretraining data of CLIP and are related to the downstream task.
This contributes auxiliary semantic knowledge to preserve the OOD generalization capabilities of CLIP.
\section{Conclusion}
In this study, we extend previous robust finetuning to a more challenging setting: preserving out-of-distribution (OOD) generalization capabilities in both domain shift and zero-shot learning during finetuning.
We argue that the diminished OOD generalization results from the overly simplified finetuning target, which provides only class information.
Consequently, we propose an Anchor-based Robust Finetuning (ARF) approach to regularize the finetuning process with auxiliary contrastive supervision.
This approach incorporates a Text-Compensated Anchor Generation module and an Image-Text Anchor Retrieval module to generate image-text-pair anchors with rich semantic information and align these anchors with contrastive loss.
Extensive experiments demonstrate the effectiveness of our approach.
\section*{Acknowledgement}

This work was supported by the National Nature Science Foundation of China under grants 62306214 and 62325111. 

{
    \small
    \bibliographystyle{ieeenat_fullname}
    \bibliography{main}
}


\end{document}